\documentclass[letterpaper, 10 pt, conference]{ieeeconf}  

\IEEEoverridecommandlockouts                              
\title{\LARGE \bf
Conflict-Based Model Predictive Control for Scalable Multi-Robot Motion Planning}

\author{Ardalan Tajbakhsh$^1$, Lorenz T. Biegler$^2$, and Aaron M. Johnson$^1$
    \thanks{$^1$ Department of Mechanical Engineering and $^2$ Department of Chemical Engineering, Carnegie Mellon University, Pittsburgh, PA, USA, \texttt{(atajbakh, lb01, amj1)@andrew.cmu.edu}}%
}
\usepackage{amsmath}
\usepackage{mathtools}
\usepackage{amsthm}
\usepackage{amssymb}
\usepackage{subfigure}
\usepackage{accents}
\usepackage{cite}
\usepackage{float}
\usepackage{romannum}
\DeclareMathOperator{\argmin}{argmin}
\usepackage{algpseudocode}
\usepackage[linesnumbered]{algorithm2e}
\usepackage{hyperref}
\hypersetup{colorlinks,allcolors=black}
\SetKwInput{KwInput}{Input}                
\SetKwInput{KwOutput}{Output}              
\hypersetup{%
  breaklinks,
  bookmarksnumbered=true,
  bookmarksopen=true,
  colorlinks=true,
  linkcolor=black,
  urlcolor=black,
  citecolor=black
}

\begin{document}
\bstctlcite{IEEEexample:BSTcontrol} 

\maketitle
\thispagestyle{empty}
\pagestyle{empty}

\begin{abstract}
This paper presents a scalable multi-robot motion planning algorithm called Conflict-Based Model Predictive Control (CB-MPC). Inspired by Conflict-Based Search (CBS), the planner leverages a modified high-level conflict tree to efficiently resolve robot-robot conflicts in the continuous space, while reasoning about each agent's kinematic and dynamic constraints and actuation limits using MPC as the low-level planner. We show that tracking high-level multi-robot plans with a vanilla MPC controller is insufficient, and results in unexpected collisions in tight navigation scenarios under realistic execution. Compared to other variations of multi-robot MPC like joint, prioritized, and distributed, we demonstrate that CB-MPC improves the executability and success rate, allows for closer robot-robot interactions, and scales better with higher numbers of robots without compromising the solution quality across a variety of environments. 
\end{abstract}
\begin{keywords}
Multi-robot motion planning, model predictive control, collision avoidance
\end{keywords}  
\section{Introduction}
In order to unlock the potential of robots in real-world applications, they often need to be deployed in numbers performing multiple tasks. These applications include picking and replenishment in warehouse fulfillment, environmental monitoring, coordinated search and rescue, material handling in hospitals, assembly operations in manufacturing, and more. Enabling such applications for multi-robot systems requires generating scalable and executable motion plans that operate in continuous time. In addition, these plans must reason about robot-robot and robot-environment constraints, be adaptable to online changes, and respect the robot kinematics, dynamics, and actuation limits.

\begin{figure}[t]
    \centering
    \subfigure[]{\includegraphics[width=0.23\textwidth]{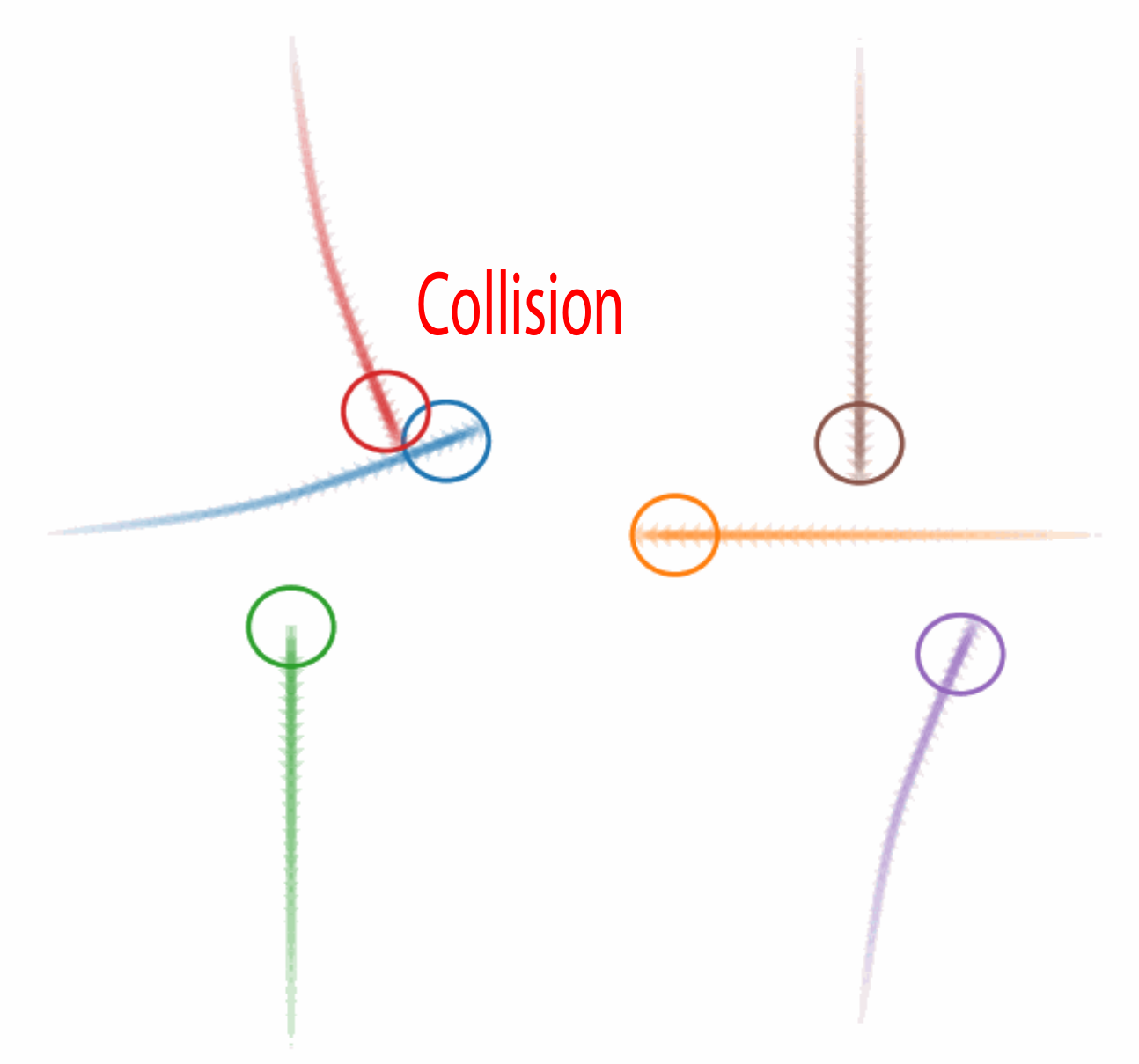}}
    \subfigure[]{\includegraphics[width=0.23\textwidth]{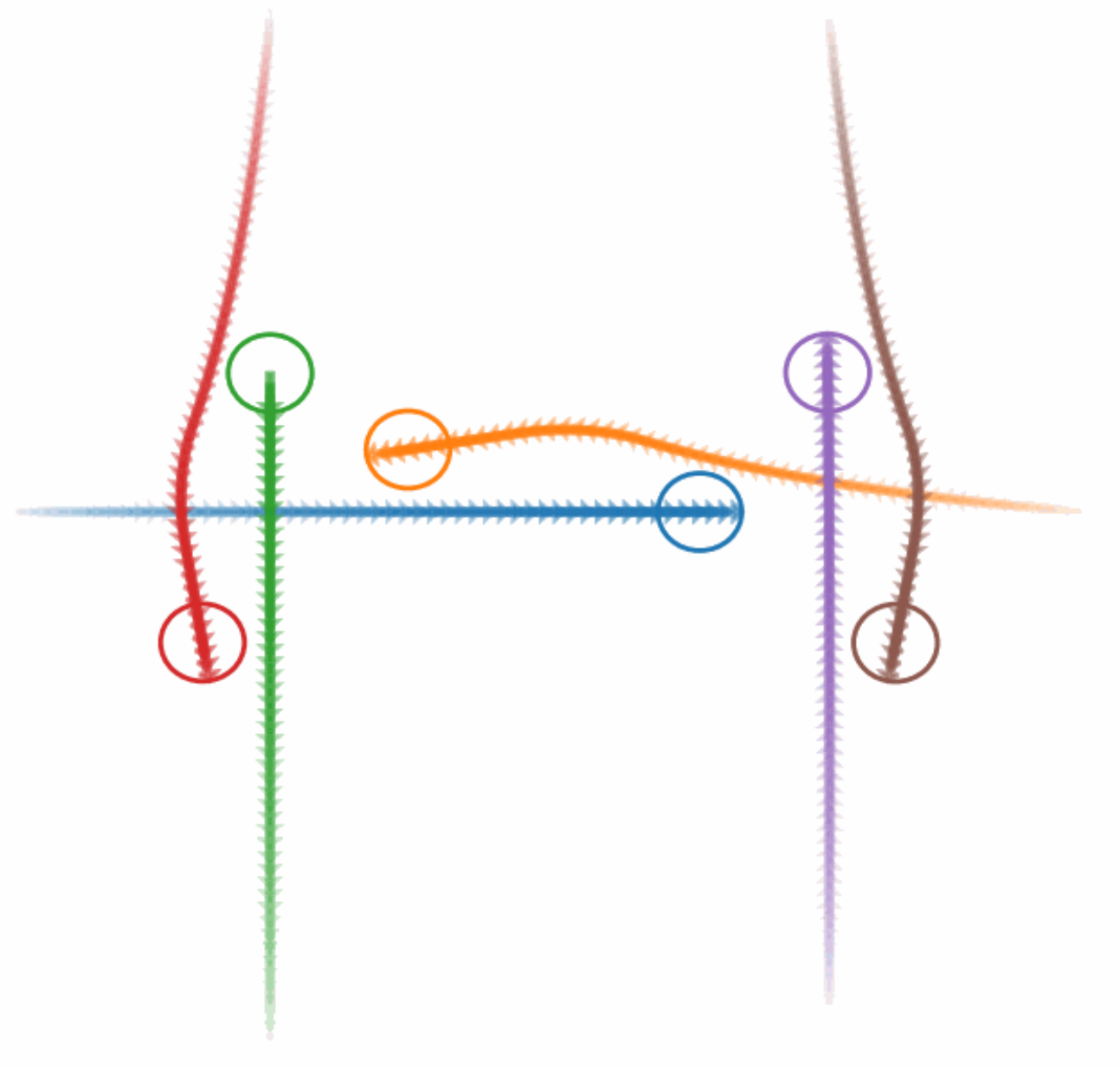}} 
    \vspace{-.5em}
    \caption{(a) Tracking a conflict-based search (CBS) generated plan with a vanilla MPC controller does not guarantee collision-free execution due to tracking error and unaccounted kinematic constraints. (b) CB-MPC allows for closer robot-robot interactions while remaining collision-free in execution.}
    \label{fig:collision}
\end{figure}

Multi-agent path finding (MAPF) algorithms have been successful at generating conflict-free position trajectories for many robots \cite{sharon2015conflict, andreychuk2022multi, boyarski2015icbs, barer2014suboptimal, li2019improved}. Although scalable, they often resort to simplifying assumptions like ignoring the kinematics, actuation limits, and continuous states and actions. As shown in Fig.~\ref{fig:collision}, this can result in infeasible executions when such plans are tracked with a controller under realistic conditions such as tracking error or actuation limits. Some approaches like  MAPF-POST \cite{honig2016multi} and action dependency graph \cite{honig2019persistent} attempt to resolve this problem by introducing an additional execution layer. However, they often result in conservative execution. Other approaches resort to reactive controllers to prevent collisions\cite{schneider2003potential, tanner2005towards, gayle2009multi, fox1997dynamic}. These methods often generate aggressive control commands and do not generalize well to tightly constrained environments with many robots.    

Multi-agent motion planning algorithms (MAMP) have also attempted to address these problems by generating dynamically-feasible plans for multiple robots while avoiding collisions \cite{vcap2013multi, solovey2015finding, kottinger2022conflict, shome2020drrt, li2021optimal, zhou2017real}. These methods often suffer from long solve times and/or poor solution quality. This is a major shortcoming as online changes in system states require an effective planner to continuously update the motion plans. To address the adaptability problem, model predictive control (MPC) based approaches generate receding-horizon state and control trajectories that respect kinematic, dynamic, and actuation limits, and avoid collisions and obstacles \cite{chen2015decoupled, luis2020online, firoozi2020distributed}. Although effective, these approaches do not scale well to large numbers of robots or tightly constrained environments due to the presence of many non-convex collision and obstacle constraints. 

Our approach differs from other multi-robot motion planning algorithms in that:
\begin{itemize}
    \item It does not include constraints from all other agents at every timestep in each optimization solve as it uses a conflict tree to resolve conflicts collaboratively. 
    \item The collisions with other robots and obstacles are efficiently resolved as constraints instead of an additional term in the cost function. This improves feasibility and solve time while guaranteeing constraint satisfaction.   
\end{itemize}

We introduce a two-level local motion planning algorithm called Conflict-Based MPC (CB-MPC) which resolves conflicts similarly to CBS combined with MPC as the low-level planner to generate collision-free and executable motion plans for multiple robots in a receding horizon fashion. These plans respect the physical limits of the system, leverage the receding horizon property of MPC to speed up the optimization by seeding with a prior solution, and resolve conflicts between agents efficiently through a CBS-like conflict tree.

This paper aims to validate the following hypotheses:
\begin{enumerate}
    \item Using CB-MPC as the local planner for receding horizon multi-robot motion planning results in a higher success rate than tracking a MAPF plan with a single robot MPC that does not include inter-robot collision constraints (Vanilla-MPC). 
    \item Compared to other MPC formulations (vanilla, joint, prioritized, and distributed), CB-MPC scales better with more robots and provides higher success rate without compromising solution quality.
\end{enumerate}
This paper is organized as follows. Sec.~\ref{sec:related} summarizes related works. Sec.~\ref{sec:preliminaries} describes the preliminaries. Sec.~\ref{sec:cb-mpc} presents the CB-MPC algorithm. Sec.~\ref{sec:results} describes the experiments and Sec.~\ref{sec:realresults} presents results in a variety of environments. Sec.~\ref{sec:conclusion} concludes the paper and notes future work.

\section{Related works}
\label{sec:related}
There have been many works that address the multi-agent path finding (MAPF) problem. The most notable example among centralized methods are conflict based search (CBS) \cite{sharon2015conflict} and its variants \cite{andreychuk2022multi, boyarski2015icbs, barer2014suboptimal, li2019improved}, which leverage a high-level conflict tree to resolve robot-robot conflicts. Another promising approach is $M^*$ \cite{wagner2011m}, which plans for each robot independently and only plans jointly for the robots that interact with each other using sub-dimensional expansion. Although scalable, these methods often require an additional execution layer like MAPF-POST \cite{honig2016multi} and action dependency graphs \cite{honig2019persistent} as they do not consider robot kinematics, actuation limits, or execution delays and they rely on discrete states and actions. Some methods resort to decentralized reactive controllers to prevent imminent collisions during execution \cite{schneider2003potential, tanner2005towards, gayle2009multi, fox1997dynamic, arul2021v}. However, they are not predictive and are prone to collisions or deadlocks in constrained environments with many agents. In \cite{li2021lifelong}, a receding horizon version of MAPF is introduced that enables continuous re-planning for all agents in dynamic environments. However, similar to other MAPF algorithms, this approach does not account for robot kinematics and actuation limits. Other methods resort to assigning priorities to agents and enforce lower priority agents to avoid the trajectories of higher priority counterparts \cite{velagapudi2010decentralized, cap2013asynchronous, bennewitz2001optimizing, li2020efficient}. This approach has been shown to perform well in environments with fewer constraints, but does not work well in highly constrained environments with many robot-robot interactions. Furthermore, it is unclear how agent priorities must be assigned in practical scenarios without sacrificing the solution quality.  

In contrast, multi-agent motion planning (MAMP) algorithms aim to generate dynamically feasible plans that robots can execute while satisfying robot-robot and robot-environment constraints.  Among sampling-based methods, \cite{vcap2013multi, solovey2015finding, kottinger2022conflict, shome2020drrt} adapt RRT and RRT* motion planners to work for multi-robot cases on discrete graphs and geometrically embedded composite roadmaps. These have been shown to outperform M* and other variants of RRT that plan over the composite state space of all agents. Recently, \cite{kottinger2022conflict} demonstrated scalable and dynamic multi-robot motion planning by leveraging the structure of CBS with a modified version of RRT to plan conflict free motion plans. Although these methods have been shown to be effective for large problems, their run-time does not allow for fast online re-planning. This can be a major drawback for hardware deployment as imperfections in execution and changes in the environment may necessitate adapting the motion plans of some agents. 

Trajectory optimization has also been applied to multi-robot problems. Some of these methods leverage sequential constraint tightening \cite{li2021optimal, zhou2017real}, since the full problem with all the constraints is often intractable. They have been shown to be effective for up to 10 robots in constrained environments. However, they are not suitable for online applications due to their large solve times. Other methods like \cite{chen2015decoupled, luis2020online, firoozi2020distributed} use model predictive control (MPC) for online trajectory generation in multi-robot systems. In these methods, robots share their predicted trajectories with the rest of the fleet and conflicts are resolved by constraining them from occupying the same locations at the same times. These methods require inter-robot collision constraints from all robots to be added at each timestep instead of resolving them collaboratively, resulting in higher solve times and potential deadlocks (especially in problems with symmetry). Our approach ensures MAPF plans are executable by introducing an efficient multi-robot MPC local planner that scales well with the number of robots using an efficient constraint splitting mechanism.  
\section{Preliminaries}
\label{sec:preliminaries}

Consider a system consisting of a set $N_a$ of nonlinear agent models, each defined as:
\begin{equation}
\label{eqn:dynamic-feasibility}
    x_i^{k+1} = f_i(x_i^k,u_i^k), \quad \forall i  \in N_a
\end{equation}
where $x_i^{k} \in X_i \subseteq R^{n}$ and $u_i^{k} \in U_i \subseteq R^m$ are the $i$th agent's states and inputs, respectively, at time $k$.

The control objective is to regulate all agents to a desired pose, $x^d_i$, while minimizing control effort and avoiding collisions with other robots and obstacles. This can be formally written as the following for every time $k$: 
\begin{align}
\label{eqn:pose regulation}
    \lim_{k\rightarrow\infty}\|x_i^d - x_i^k\| &\leq \epsilon_g  &\forall& i \in N_a\\
\label{eqn:inter-agent separation}
    \|p_j^k - p_i^k\| &\geq D + \delta_r + \epsilon_r, &\forall& i, j \in N_a \\
\label{eqn:obstacle avoidance}
    \|p^o_p - p_i^k\| &\geq \frac{D}{2} + \delta_o + \epsilon_o, &\forall& i \in N_a, p \in N_o
\end{align}
where, $\epsilon_g$ is the goal tolerance, $D$ represents the robot footprint diameter, $p_i^k$ is the position component of $x_i^k$, the robot state, $p^o_p$ is the position of the static obstacle $p$, $N_o$ is the set of static obstacles in the environment, $\epsilon_r$, $\delta_r$, $\epsilon_o$, and $\delta_o$ are the safety margins and slack variables for robot and obstacle constraints, respectively. Note that we only consider circular and polyhedral obstacles in the presented experiments, but maintain that most irregular shaped obstacles can be safely approximated with collections of these two primitives.

\begin{figure*}[t!]
  \centering
  \subfigure(a){\fbox{\includegraphics[width=0.2\textwidth]{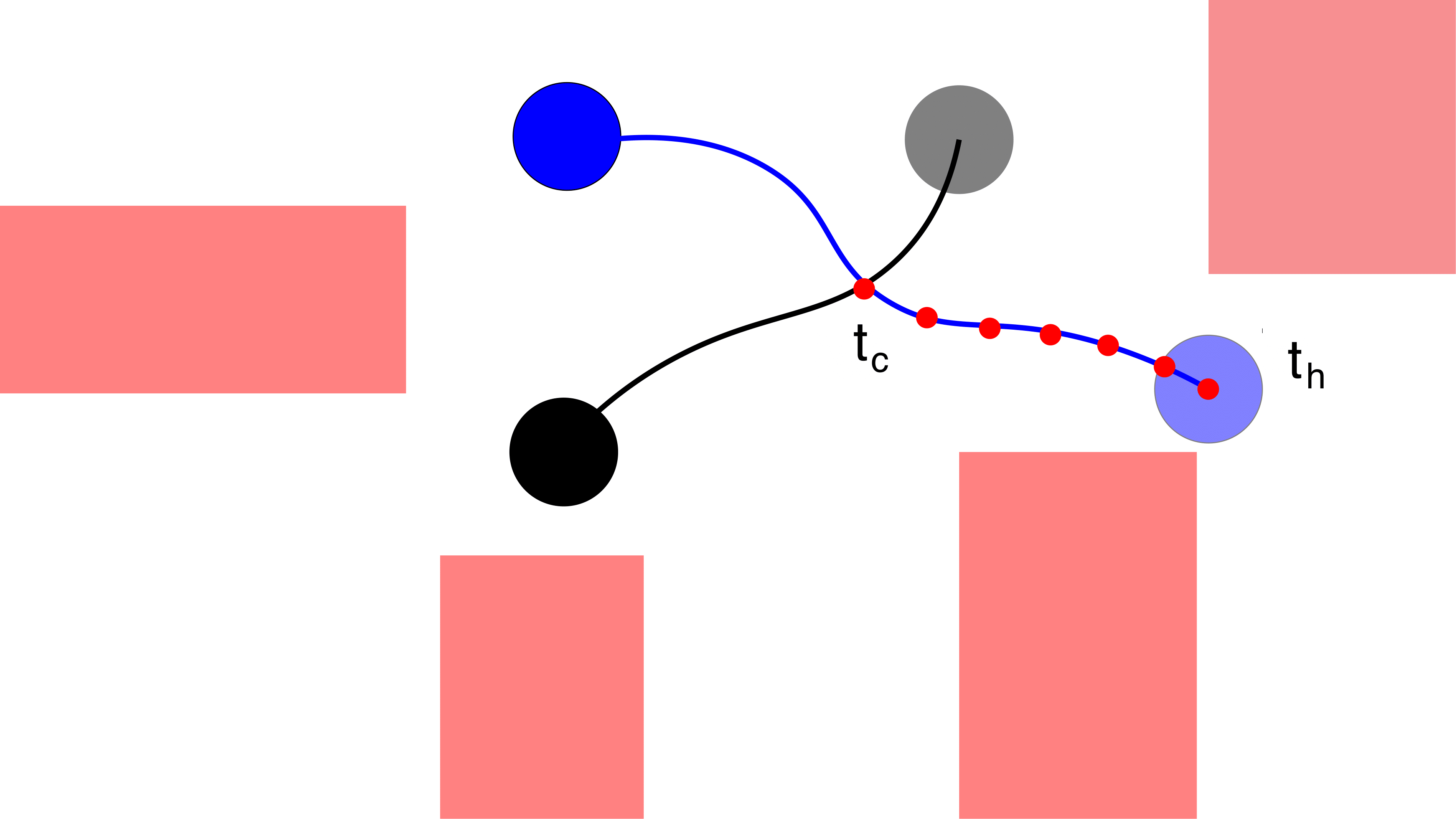}}}
  \subfigure(b){\fbox{\includegraphics[width=0.2\textwidth]{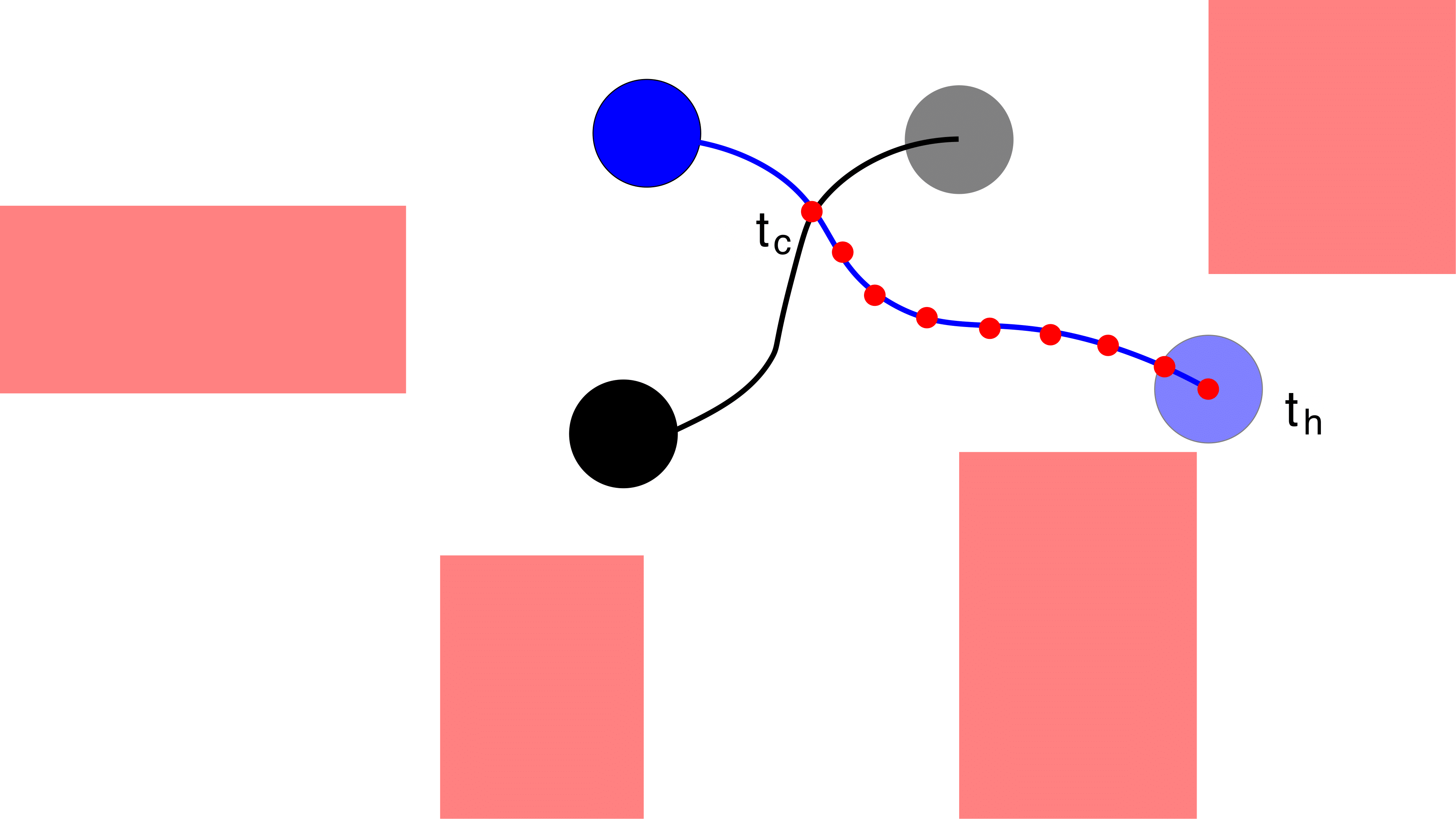}}}
  \subfigure(c){\fbox{\includegraphics[width=0.2\textwidth]{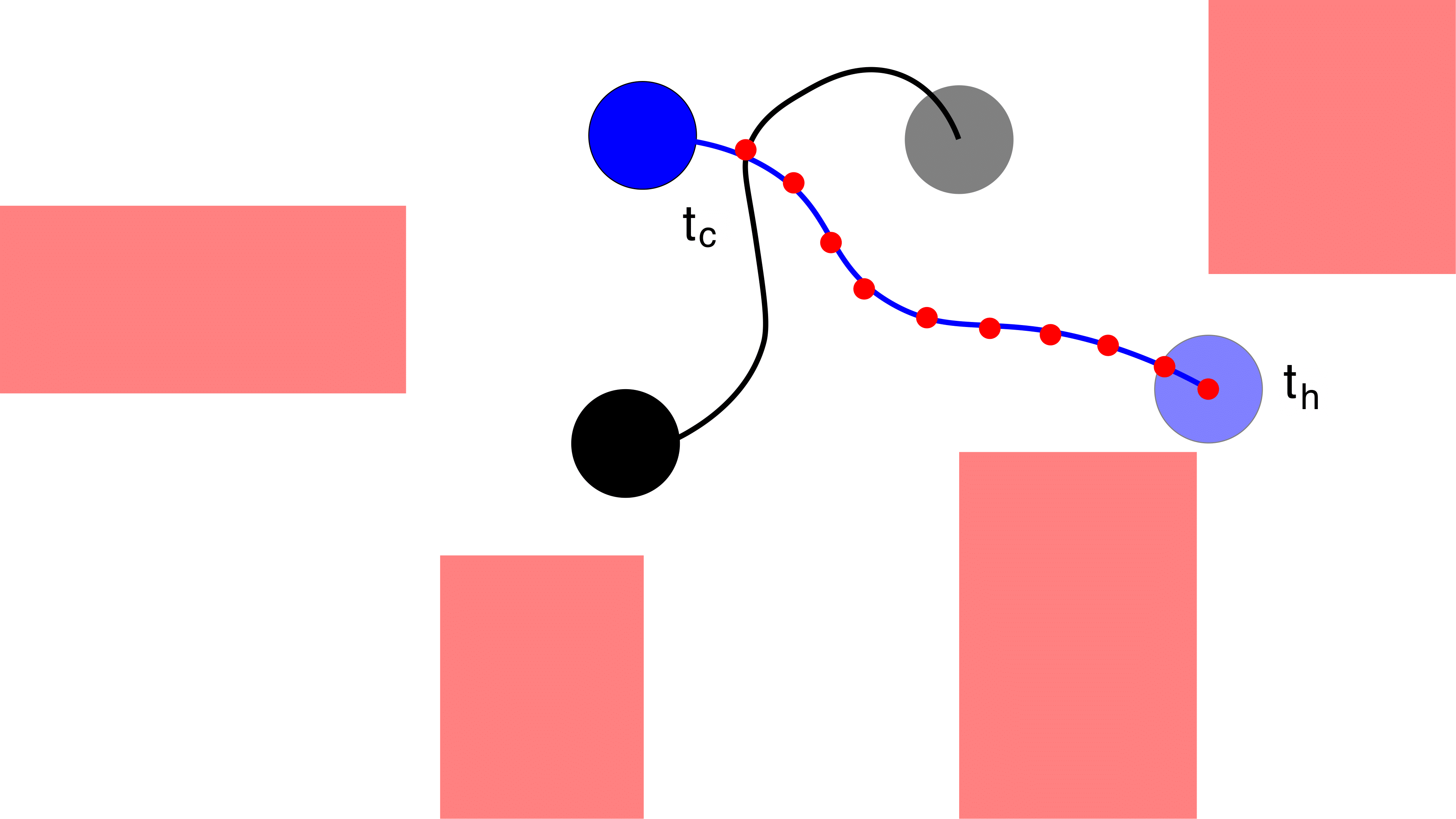}}}
  \subfigure(d){\fbox{\includegraphics[width=0.2\textwidth]{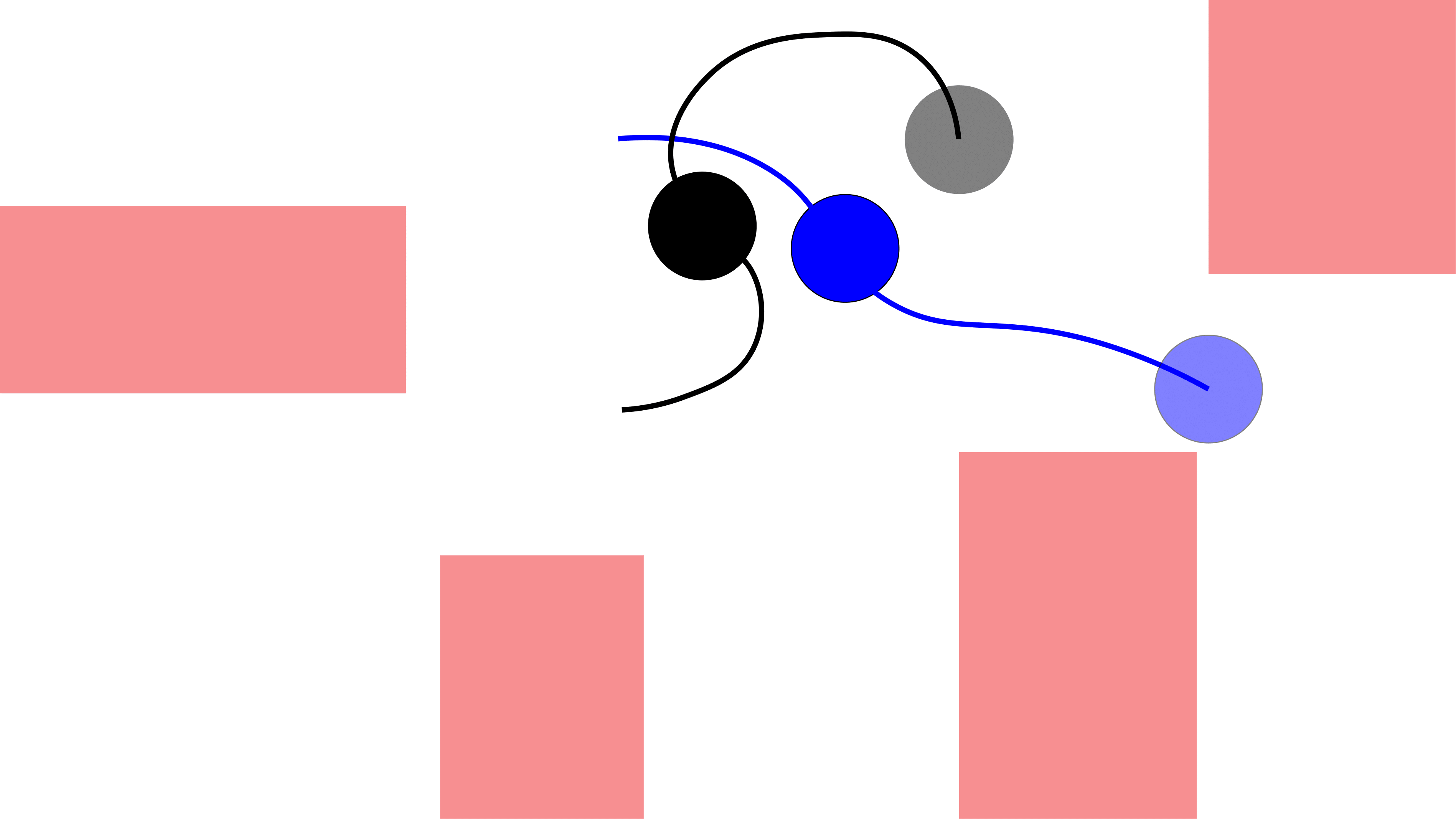}}}
  \caption{(a) Agents black and blue are tasked to go from their start poses (shown in solid colors) to their goal poses (shown in transparent colors). For a given planning horizon $t_h$, the initial predicted MPC trajectory of the agents has a collision at $t_c$. Collision constraints are added at each timestep between [$t_c$, $t_h$] (shown in red dots) for the black agent since it has a lower total cost. (b,c) The updated MPC solution resolves the prior constraints, but results in a new collision. (d) Collision-free trajectories are generated after collisions are resolved iteratively.}
  \vspace{-1em}
\label{fig:collision description}
\end{figure*}

\subsection{Conflict Based Search (CBS)}
CBS is a centralized, complete, and optimal multi-agent path finding algorithm that relies on collaborative conflict resolution to efficiently generate conflict-free paths for multiple agents simultaneously \cite{sharon2015conflict}. In particular, CBS uses a low-level path planner (for example $A^*$ \cite{duchovn2014path}) to plan single-robot plans. It then checks those generated plans for vertex or edge conflicts (i.e.\ robots occupying same locations at the same time, or traversing the same edge at the same time). Conflicts are resolved by invoking the low-level planner with the additional constraints. This process is repeated until a conflict-free plan is found for all the agents. The key idea with CBS is splitting constraints between the agents and adding them incrementally. This property allows for better scalability as the problem size grows with the number of conflicts and not the number of agents. We borrow a similar conflict resolution mechanism from CBS with modifications to the conflict definition (defined in \ref{sec:conflict definition}) and the application of those conflicts as constraints in receding horizon planning.  

\subsection{Model Predictive Control Planner (MPC)}
MPC is a receding horizon optimization framework that is commonly used for solving constrained optimal control problems. The optimization problem in MPC is often solved under dynamic, state, and input constraints to ensure solutions can be closely executed on hardware. In the case of MAMP, additional nonlinear constraints from other robots and obstacles are present, which make the optimization non-convex. However, most modern non-linear optimization solvers can effectively solve these problems using either sequential quadratic programming (SQP) or interior point (IP) methods. The MPC problem for each robot is as follows:
\begin{align}
\label{eqn:mpc-objective}
    \argmin _{u_i,x_i,\delta_r,\delta_o}J_i = \sum_{l=k}^{k+N-1}((x_i^l-r_i^l)^TQ(x_i^l-r_i^l) \nonumber \\ + u_i^{l,T}Ru_i^{l}) + x_i^{(k+N),T}Px_i^{k+N} + k_r\delta_r + k_o\delta_o
\end{align}
subject to (\ref{eqn:dynamic-feasibility}), (\ref{eqn:inter-agent separation}), (\ref{eqn:obstacle avoidance}),
\begin{align}
    \label{eqn:initial condition}
    x_i^k &\in X_0 
   \qquad & x^{k+N}_i &\in X_f \\
    \label{eqn:state feasibility}
    x_i^l &\in X_{feasible} 
    \qquad & u_i^l &\in U_{feasible} \\
    \label{eqn:slack feasibility}
    \delta_r, \delta_o &\geq 0
\end{align}
for all $l$ over the horizon $k$ to $k+N$ and all robots $i \in N_a$.
In this formulation, $u_i = [u_i^k, u_i^{k+1},...,u_i^{k+N-1}]$, $x_i = [x_i^k, x_i^{k+1},...,x_i^{k+N}]$, and $r_i = [r_i^k, r_k^{k+1},...,r_i^{k+N}]$ denote the sequence of control inputs, states, and reference states. $X_0$ and $X_f$ denote the set of initial and final conditions for all agents. $X_{feasible}$ and $U_{feasible}$ represent the set of feasible states and control inputs. The objective function (\ref{eqn:mpc-objective}) penalizes reference tracking error, control input magnitude, and terminal cost. $Q$, $R$, and $P$ are positive semi-definite weighting matrices. (\ref{eqn:dynamic-feasibility}), (\ref{eqn:inter-agent separation}), and (\ref{eqn:obstacle avoidance}) represent the kinematic and dynamic feasibility, inter-robot collision, and obstacle constraints, respectively. (\ref{eqn:initial condition}) enforces the initial and terminal conditions for all agents. (\ref{eqn:state feasibility}) and (\ref{eqn:slack feasibility}) ensure state, control, and slack variable feasibility. Note that using the $l_1$ relaxation of the collision constraints is critical for convergence in tight navigation problems. When a reference is needed, we use CBS as the high-level path planner to generate $r$, since it reasons about conflicts between agents and acts as an effective heuristic.

\section{Conflict-Based Model Predictive Control}\label{sec:cb-mpc}
\subsection{Conflicts in CB-MPC}
\label{sec:conflict definition}
As shown in Fig. \ref{fig:collision description}, for a given planning horizon that ends at time $t_h$, conflicts can arise between pairs of agents, or agents and obstacles. For agent-agent conflicts between agents $a_i$ and $a_j$, the conflicts are defined as a tuple ($a_i$, $a_j$, [$t_c$, $t_h$]), where $t_c$ is the predicted time of collision when \eqref{eqn:inter-agent separation} is violated. Note that all agent-obstacle constraints are added for a given agent as defined in \eqref{eqn:obstacle avoidance} at each timestep. Compared to CBS, conflicts in CB-MPC are defined over location-time ranges instead of location-time pairs. We found that adding individual constraints at each timestep shifts the collision to the immediate future, causing unnecessary extra solves.

\subsection{CB-MPC Algorithm}
CB-MPC combines the benefits of MPC and CBS in a unified motion planning framework to allow for receding horizon multi-robot motion planning. At each timestep, the algorithm, summarized in Algorithm \ref{proc:alg name}, takes as input the tuple $M = (N_o, X_i, X_f, N)$ that summarizes all the problem specific variables. Each robot plans its motion for a given planning horizon using MPC with all the constraints excluding (\ref{eqn:inter-agent separation}) and (\ref{eqn:obstacle avoidance}) (lines 2-4 in Algorithm \ref{proc:alg name}).
The resulting cost $J_n$ computed by the sum of individual agent costs ($SIC$), which is defined as the sum of the current trajectory length $J_c$, and the cost-to-go $J_{f}$, for all agents (line 5). $J_f$ in our case is the Euclidean distance from the final position of the horizon to the goal. Note that in cases where agents have different capabilities (e.g.\ drones and ground vehicles), this cost can also include the agent's velocity and acceleration.
The trajectories and the total cost are stored in $R.solution$ and $R.cost$ as a node, and the node is added to an open list that is sorted by cost in an increasing order (lines 6-7). 

The lowest cost node from the open list is retrieved and removed (line 9). The trajectories stored in this node are checked for collisions for the involved pair of agents at each timestep in the planning horizon (line 10). If these trajectories are collision-free with respect to (\ref{eqn:inter-agent separation}) and (\ref{eqn:obstacle avoidance}), the solution is returned (lines 11-13). The first input is then applied for each robot, and the state is propagated based on the dynamics. Otherwise, the first conflict (smallest $t_c$) is retrieved and a new node is created for each agent involved in that conflict (lines 14-16). For a given planning horizon $t_h$, the collision constraints are then applied at each time between [$t_c$, $t_h$] (refer to Fig.~\ref{fig:collision description}). 

The application of constraints for all timesteps after $t_c$ is done because constraining a single timestep will likely shift the collision to the immediate future in the receding horizon setting. Furthermore, in most practical scenarios, collisions will appear from the end of the horizon, which results in most optimization solves containing collision constraints for a small subset of the planning horizon. Next, the MPC optimization runs for the two conflicting agents with the updated constraints, the solution of the node are updated, and the nodes are added to the open list (lines 17-20). This process continues until a conflict-free plan is found.

\begin{algorithm}[t]
\caption{Conflict-based MPC}
 \label{proc:alg name}
\DontPrintSemicolon
  \KwInput{$M$}
  \KwOutput{Collision-free trajectories $P
  .solution$}
   $R$ $\leftarrow$ new Node \\
  \For{each agent $a_i$} {
   $R.solution(a_i)$ = $MPC(M, a_i,\{\})$ 
  } 
   $R.cost$ = $SIC(R.solution)$ \\
  O = \{\} \\
  Insert $R$ in O \\ 
 \While {True:} {
    $P$ $\leftarrow$ lowest cost node from O \\
    check $P$ for conflicts \\
 \If {$P$ has no conflicts:} { 
    \textbf{return} $P.solution$
 }
 $C$ $\leftarrow$ first conflict ($a_i$, $a_j$, $[t_{c_1}, t_{c_2}]$) \\
 \For{each agent $a_i$ in $C$} {
    $A$ $\leftarrow$ new Node \\
    A.constraints = $P.constraints$ + $(a_i,a_j,[t_{c_1}, t_{c_2}])$ \\
    $A.solution(a_i)$ = $MPC(M, a_i, A.constraints)$ $\forall a_j \neq a_i$\\
    $A.cost$ = $SIC(A.solution)$ \\
    Insert $A$ to O
 }
 }
\end{algorithm}
Note that similar to any other NMPC-based planner, CB-MPC does not provide feasibility or global optimality guarantees given an arbitrary feasible problem. However, we empirically show that CB-MPC outperforms other benchmarks in terms of success rate and scalability across many tight navigation scenarios in different environments. In practice, CB-MPC implementation can either be done centralized or distributed with full inter-agent communication. Furthermore, the algorithm as defined requires synchronous communication among agents. This requirement can be easily relaxed since robots can use the last received plan of other neighbors even if it is delayed. In such case, the safety margin $\epsilon_r$ can be increased proportional to the time since the last received communication, similar to \cite{ferranti2022distributed}.

\begin{table}[t] \setlength\tabcolsep{4 pt} 
\caption{Summary of hyper-parameters used in the experiments}
\begin{center}
\vspace{-1em}
\begin{tabular}{c | c | c}
 Quantity & Description & Value \\
 \hline
 Q & Reference tracking penalty &  $diag[12.5,12.5]$ \\
 R & Control effort penalty & $diag[12.5, 0.05]$ \\
 P & Goal tracking penalty & $diag[12.5, 12.5]$ \\
 N & MPC horizon & 60 \\ 
 D & Robot footprint [$m$] & 0.3 \\
 $\epsilon_g$ & Goal tolerance [$m$] & 0.2 \\
 $\epsilon_r$ & Robot-robot collision tolerance [$m$] & 0.05 \\
 $\epsilon_o$ & Robot-obstacle collision tolerance [$m$] & 0.05 \\
 $k_r$ & Coefficient for $\epsilon_r$ & $10^6$ \\
 $k_o$ & Coefficient for $\epsilon_o$ & $10^6$
\end{tabular}
 \label{tab:hyperparams}
\end{center}
\end{table}

\section{Experiments}
\label{sec:results}
We conducted a series of experiments to validate the efficacy of the CB-MPC algorithm and compare its performance against Vanilla-MPC, Joint-MPC, prioritized (Pr-MPC), and distributed (D-MPC) methods for different planning horizons ($N$), number of robots ($N_{rob}$), and a variety of environments. 

\subsection{Baseline Comparisons}
The Joint-MPC refers to the centralized problem, which is structurally similar to \cite{riegger2016centralized}. In this case, the optimization is solved over the joint state space of all robots, and all obstacle and collision constraints are present in the problem when passed to the solver. Pr-MPC is implemented similar to \cite{li2020efficient} with random priority assignment between the agents. In this case, the lower priority robots must avoid the predicted trajectory of the higher priority robots. D-MPC is another variation of multi-robot MPC \cite{luis2020online}, which detects conflicts on-demand similar to CB-MPC. However, conflicts are resolved by all the agents involved and not collaboratively due to the absence of a coordinator. The Vanilla-MPC is the standard single-agent MPC that does not include any inter-agent collision constraints and closely tracks the CBS reference. Note that for all the baseline comparisons, we only implement the core MPC algorithms. In particular, we do not perform any additional convexification of the collision constraints, and, when using a reference, utilize CBS as the high-level planner to keep the comparisons uniform.

\subsection{Experimental Setup}
Experiments are performed using a nonlinear kinematic unicycle model \cite{malu2014kinematics} across three different environments: a narrow environment, an open environment, and a randomized environment with obstacles. All experiments are implemented in Python and run on a 6-core Intel core i7 @ 3.7 GHz with 64 GB of RAM. The optimization problems are modeled in Casadi \cite{andersson2019casadi} and solved using IPOPT \cite{biegler2009large} with warm-starting enabled. Additionally, we implemented parallelization for Vanilla-MPC, CB-MPC and D-MPC where appropriate. Table \ref{tab:hyperparams} summarizes the default hyper-parameters used throughout the experiments.

A trial is considered successful if all robots reach their goal state without any collisions. If an optimization solve returns an infeasible solution or a trial exceeds 500 timesteps, we mark it as a failure. In addition, we perform collision checking at each timestep of the final executed solution. We report success rate, makespan (i.e.\ the time required for all robots to reach their goals), average solve time per robot ($T_{avg}$), max solve times per fleet ($T_{max}$), and average number of constraints added per timestep per robot ($C_{avg}$).

\begin{table}[t] \setlength\tabcolsep{4 pt} 
\caption{Benchmark results for the narrow environment (Note that T and C denote failures due to timeout and collision)}
\begin{center}
\begin{tabular}{ c| c c c c c c} 
 Algorithm & $N$ & Makespan & $T_{avg}$ & $T_{max}$ & $C_{avg}$ & Failure \\
 \hline
 CB-MPC & 10 & - & - & - & - & \textcolor{red}{T} \\
 Pr-MPC & 10 & - & - & - & - & \textcolor{red}{C} \\
 D-MPC & 10 & - & - & - & - & \textcolor{red}{T}\\
 Vanilla-MPC & 10 & - & - & - & - & \textcolor{red}{C}\\
 Joint-MPC & 10 & \textbf{7.85} & \textbf{0.09} & \textbf{0.32} & \textbf{20} & - \\ 
 \hline
  CB-MPC & 20 & \textbf{8.35} & \textbf{0.08} & \textbf{0.74} & \textbf{1.35} & - \\
   Pr-MPC & 20 & - & - & - & - & \textcolor{red}{C}\\
 D-MPC & 20 & - & - & - & - & \textcolor{red}{T}\\
 Vanilla-MPC & 20 & - & - & - & - & \textcolor{red}{C}\\
 Joint-MPC & 20 & 8.45 & 0.38 & 1.10 & 40 & -
\end{tabular}
\end{center}
\label{tab:experiment_1}
\end{table}

\begin{figure}[t]
    \centering
    \vspace{-1em}
    \includegraphics[width=0.4\textwidth]{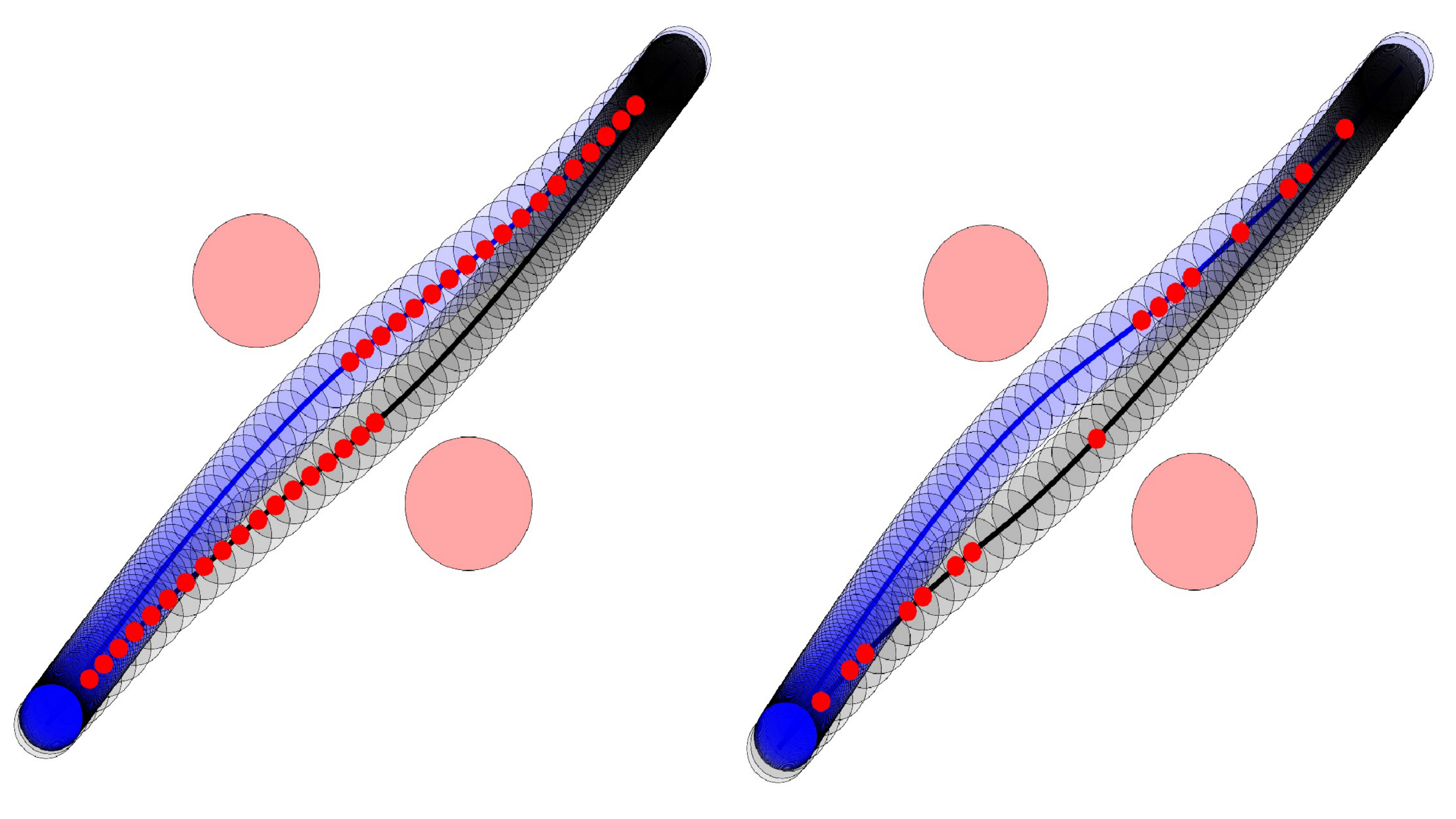}
    \vspace{-1em}
    \caption{Comparison between the Joint-MPC (left) and CB-MPC (right) in the narrow environment. CB-MPC is able to get similar quality of solution with the makespan of 8.35 compared to 8.45 of the Joint-MPC at a significantly lower computation cost. Red dots denote the active constraints.}
    \label{fig:narrow_env}
\end{figure}
\section{Results}
\label{sec:realresults}
\subsection{Narrow Environment}
The narrow environment, as shown in Fig. \ref{fig:narrow_env}, is a challenging benchmark that requires two robots to swap positions through a narrow corridor in presence of tight obstacle and robot-robot collision constraints (a four-robot scenario for this environment is also presented in the video attachment).

As shown in Table \ref{tab:experiment_1}, only Joint-MPC and CB-MPC are able to complete the task with a planning horizon of $N=20$. Note that CB-MPC is able to do so with a similar makespan at approximately a quarter of the $T_{avg}$ of the Joint-MPC. This is due to CB-MPC solving smaller optimization problems with fewer robot-robot constraints as demonstrated by the active constraints for the entire trajectory in red in Fig.~\ref{fig:narrow_env}. The computational cost of the Joint-MPC comes with the advantage of fewer infeasibilities. This is enabled by the additional degrees of freedom in the optimization. In particular, at $N$ = 10, only Joint-MPC finds a feasible solution. From this observation, we can conclude that the constraints of this problem require both robots to simultaneously modify their trajectories with shorter planning horizons. Comparing CB-MPC with other MPC variations, note that the collaborative conflict splitting mechanism enables CB-MPC to generate feasible motions through the narrow corridor where other approaches become infeasible or get stuck in a deadlock. This supports \textbf{hypothesis 2}, since CB-MPC outperforms other multi-robot MPC variations in this challenging environment. 

\begin{table}[t] \setlength\tabcolsep{4 pt} 
\caption{Benchmark results for the open environment}
\label{tab:open_env}
\begin{center}
\vspace{-1em}
\begin{tabular}{ c|c| c c c c c c} 
 $N_{rob}$ & Algorithm & Makespan & $T_{avg}$ & $T_{max}$ & $C_{avg}$ & Failure\\
 \hline
 4 & CB-MPC &  \textbf{6.30} & \textbf{0.13} &  \textbf{1.40} & \textbf{9.90} & - \\
  & Pr-MPC & 6.50 & 0.21 & 1.52 & 22.5 & - \\
 & D-MPC & - & - & - & - & \textcolor{red}{T} \\
 & Vanilla-MPC & 7.45 &  0.23 &  1.09 & - & -\\
 & Joint-MPC & - & - & - & - & \textcolor{red}{T}\\
 \hline
 6 & CB-MPC & \textbf{7.80} & \textbf{0.10} & \textbf{1.66} & \textbf{7.99} & - \\
  & Pr-MPC & 8.90 & 0.32 & 3.17 & 25.00 & - \\
   & Vanilla-MPC & - & - & - & - & \textcolor{red}{C}\\
 \hline
 8 & CB-MPC &  10.90 & \textbf{0.12} & \textbf{3.50} & \textbf{11.83} & - \\
  & Pr-MPC & \textbf{9.15} & 0.37 & 4.47 & 26.25 & - \\

 \hline
 12 & CB-MPC & \textbf{9.20} & \textbf{0.14} & \textbf{7.15} & \textbf{16.74} & - \\
  & Pr-MPC & - & - & - & - & \textcolor{red}{C} 
\end{tabular}
\end{center}
\end{table}

\begin{figure}[t]
    \centering
    \vspace{-1em}
    \subfigure[]{\includegraphics[width=0.23\textwidth]{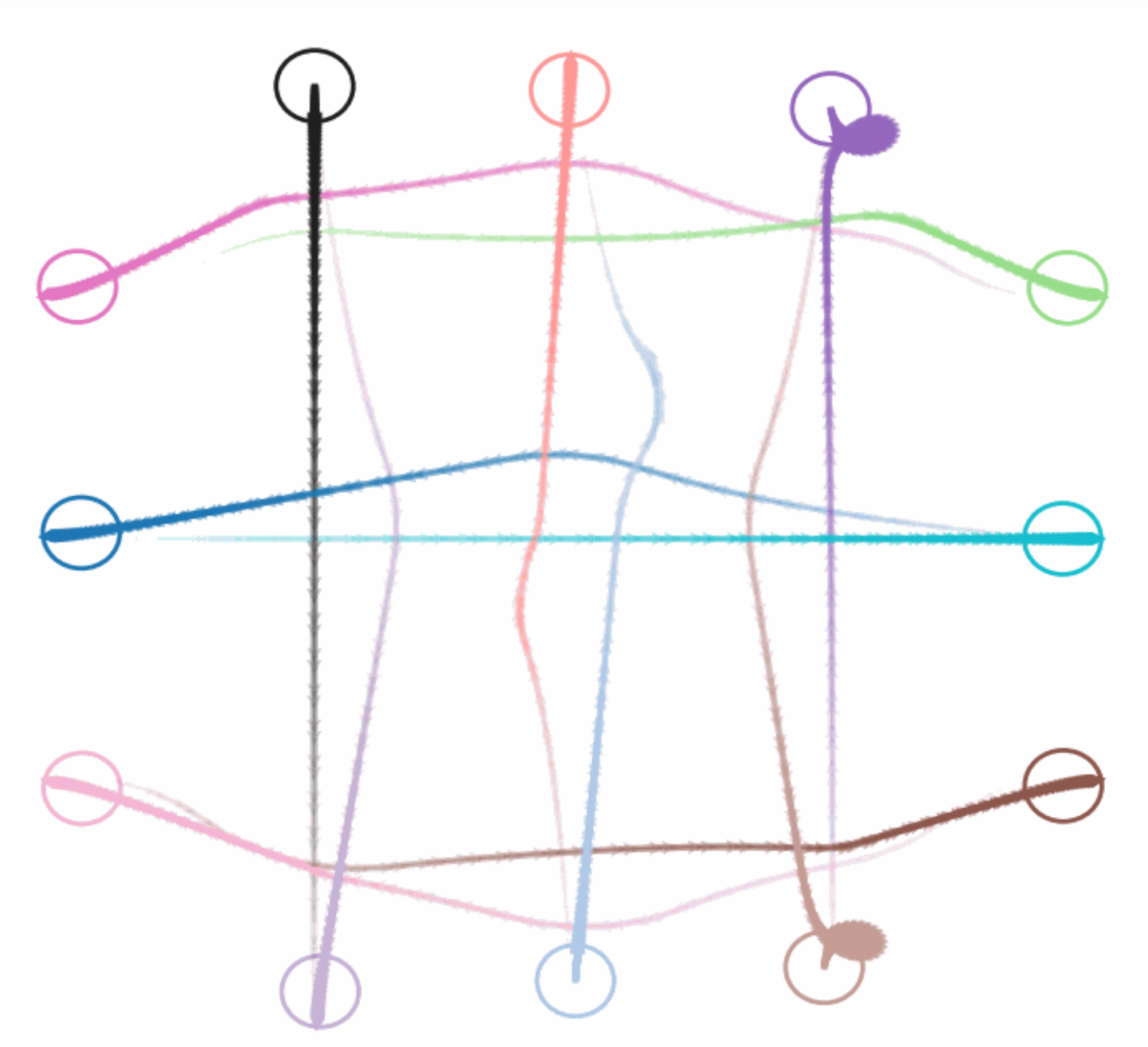}}
    \subfigure[]{\includegraphics[width=0.23\textwidth]{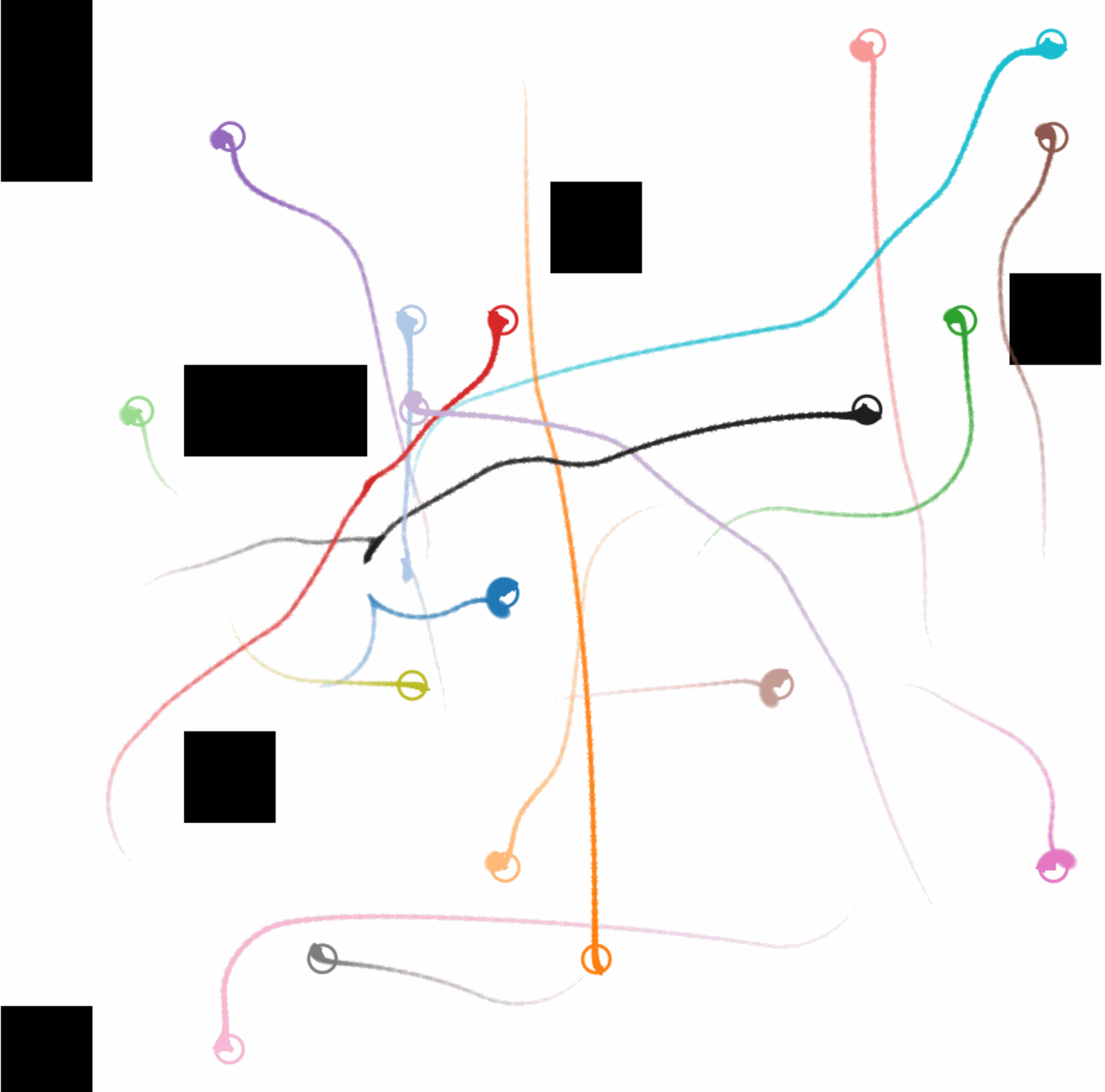}} 
    \vspace{-.5em}
    \caption{Only CB-MPC can solve: (a) the 12-robot problem in the open environment (b) the 18-robot problem in the randomized environment.}
    \label{fig:open and cluttered}
\end{figure}

\subsection{Open Environment}
The open environment demonstrates a particularly difficult case in terms of the number of robot-robot interactions. Robots forming a square are tasked to switch positions with the robot across from them, while avoiding all other robots, Fig. \ref{fig:collision} and Fig.~\ref{fig:open and cluttered}(a). 
Note that except for the Vanilla-MPC, we do not use a reference trajectory for the other algorithms in order to increase the number of robot-robot interactions.

\begin{figure*}[t]
  \centering
  \subfigure{
  \includegraphics[width=0.25\textwidth]{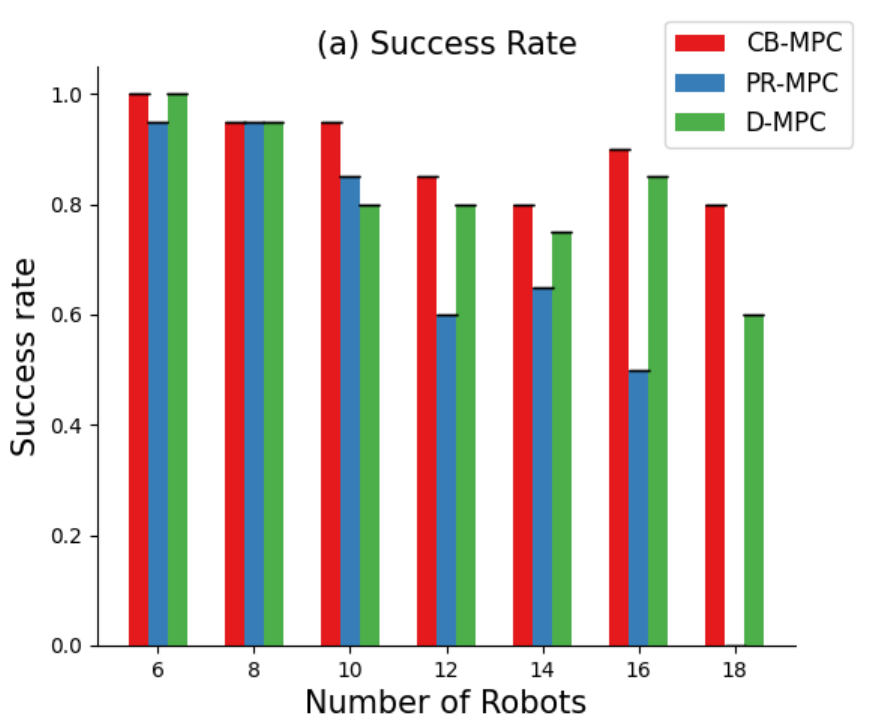}} 
  \subfigure{\includegraphics[width=0.24\textwidth]{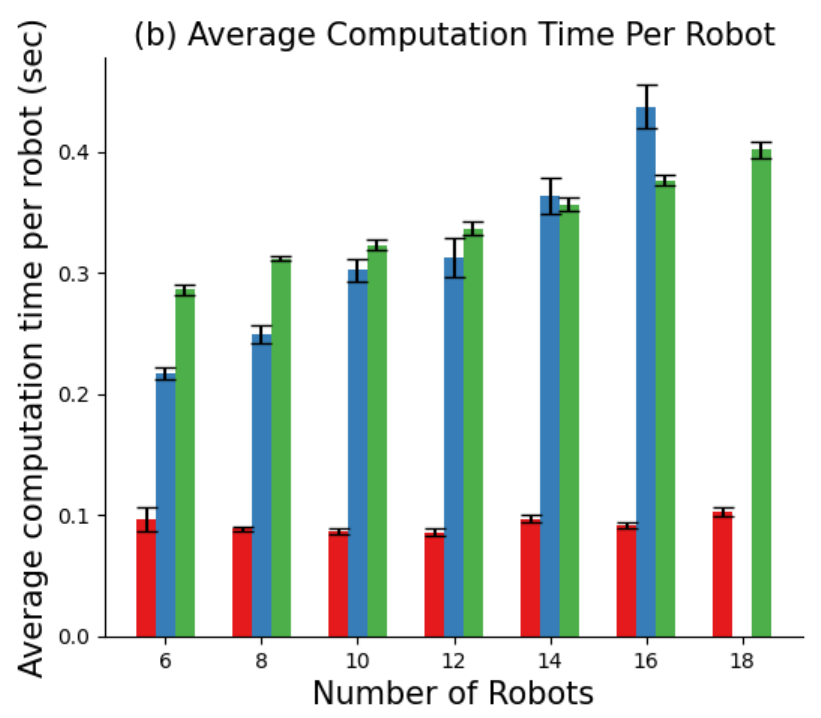}} 
  \subfigure{\includegraphics[width=0.24\textwidth]{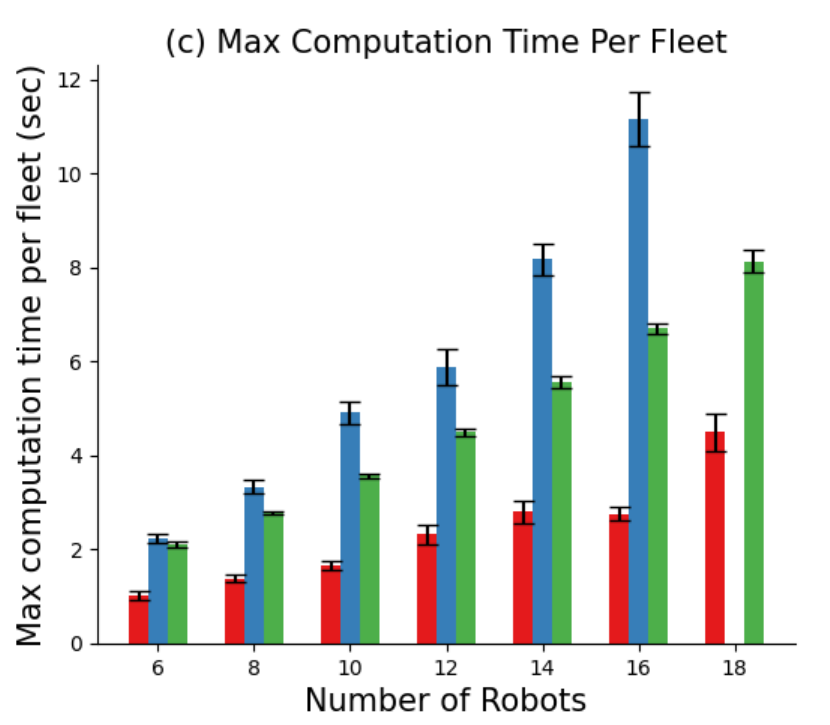}}
  \subfigure{\includegraphics[width=0.24\textwidth]{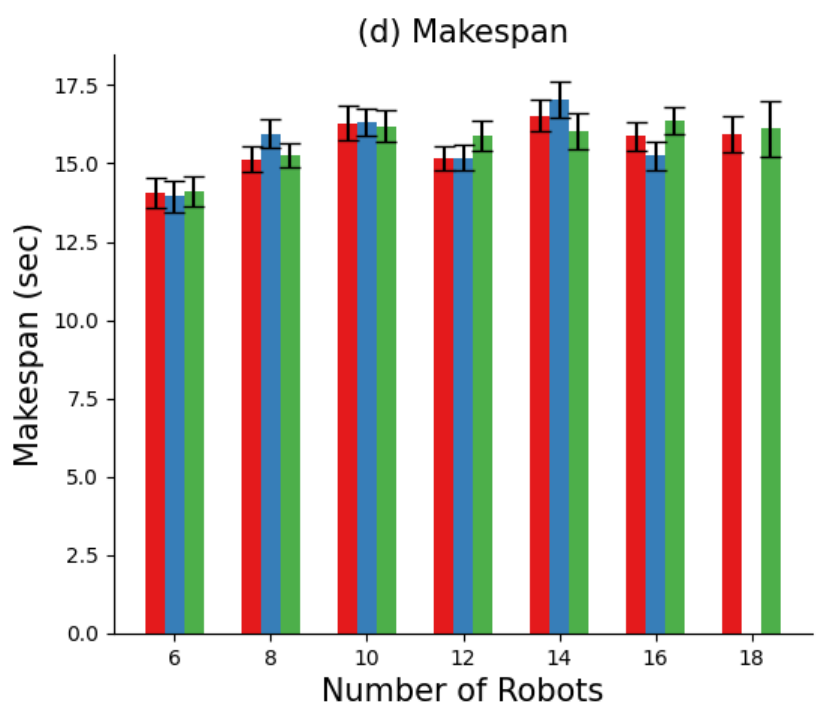}}
  \vspace{-2em}
  \caption{Randomized environment results. (a) Success rate (higher is better) (b) Average solve time per robot (lower is better) (c) Max solve time per fleet (lower is better) (d) Makespan (lower is better). CB-MPC results in higher success rate in almost all cases and significantly better average and max solve times compared to D-MPC and Pr-MPC, with similar makespan.}
  \vspace{-1em}
\label{fig:cluttered_env}
\end{figure*}

As shown in Fig. \ref{fig:collision}(a), Vanilla-MPC results in collision during execution for cases with more than four robots, despite tracking the collision-free CBS plan. This is due to the kinematics and actuation limits of the robots that are not accounted for and resolving the conflicts in discrete-time under the assumption of ideal execution. As a result, slight mismatches in agents' timing tracking the reference waypoints can result in unsafe behaviour. This issue can still exist even if the reference is dynamically feasible due to other execution imperfections. D-MPC was unable to solve any of the instances and resulted in deadlocks due to lack of coordination between the robots. Pr-MPC also resulted in collisions with more than eight robots as the excessive constraints added to lower priority robots forced the solver to violate the soft collision constraints. CB-MPC outperforms other baselines in terms of solution quality (i.e.\ makespan) and is the only algorithm that is able to solve all the cases up to 12 robots (refer to Table \ref{tab:open_env}). Notably, the conflict tree in the 12 robot scenario shown in Fig. \ref{fig:open and cluttered}(a) reached a maximum depth of 19 to resolve all the conflicts. 

It is important to note that CB-MPC maintains better $T_{avg}$ and $T_{max}$ compared to Pr-MPC across all cases, and it scales better as the number of robots increases (the average time per robot is almost identical from 4 to 12 robots). This is mainly due to the fact that CB-MPC handles inter-robot collision constraints more effectively by splitting them between the involved robots and resolving them incrementally. The incremental addition of constraints results in significantly smaller $C_{avg}$ than Pr-MPC, since CB-MPC only uses the minimal set of constraints necessary to solve a given problem. In comparison, Pr-MPC penalizes lower priority robots significantly more, since they have to avoid the predicted trajectories of all higher priority robots. This results in lower priority robots solving harder optimization problems, resulting in compromised solution quality or failures due to timeouts.

To summarize, we demonstrate that CB-MPC outperforms other baselines on this particularly difficult scenario \textbf{(hypothesis 2)}. Furthermore, we showed that CB-MPC guarantees collision-free execution as opposed to tracking a MAPF plan with single robot vanilla MPC \textbf{(hypothesis 1)}.

\subsection{Randomized Environment}
The randomized environment presents a set of randomly generated realistic navigation scenarios, Fig.~\ref{fig:open and cluttered}(b). We exclude the Vanilla-MPC and the Joint-MPC from these trials due to the fundamental issues of executability and scalability discussed previously. Each environment is a randomly generated 12x12 grid map with 8 randomly placed obstacles. The start and end positions of robots are randomly selected for each scenario such that the start/goal positions are within the free space and are not overlapping. We have experimented with 6, 8, 10, 12, 14, 16, and 18 robots and performed 20 trials per scenario, resulting in a total of 140 scenarios. 

As shown in Fig. \ref{fig:cluttered_env}(a), CB-MPC provides at least as good or better success rate compared to D-MPC and Pr-MPC across all the scenarios. This advantage is due to splitting the non-convex collision constraints between the involved agents and resolving them incrementally. In comparison, D-MPC and Pr-MPC suffer from either collision constraint violation due to solving overly constrained problems, or deadlocks resulting from lack of coordination among agents. This is exacerbated as the number of robots, and consequently the number of robot-robot interactions, increases.   

Furthermore, CB-MPC is able to maintain this success rate with a significantly better $T_{avg}$ and $T_{max}$ (Fig.~\ref{fig:cluttered_env}(b) and (c)). In particular, the $T_{avg}$ remains mostly constant as the number of robots increases, thus demonstrating superior scalability compared other benchmarks. While CB-MPC requires resolving multiple optimization problems when dealing with conflicts, each subsequent solve is warm-started with the previous solution and only has a single additional constraint set, which reduces the solve times significantly. This would mean that in most timesteps CB-MPC is essentially solving a set of single-agent MPC problems in parallel when there are no inter-agent conflicts, while taking advantage of the computational efficiency of its conflict resolution mechanism in timesteps where such conflicts are present. Finally, as shown in Fig. \ref{fig:cluttered_env}(d), CB-MPC maintains similar makespan compared to other benchmarks. 

To summarize, this experiment provides additional evidence for \textbf{hypothesis 2}, as CB-MPC is shown to out-perform other baselines in terms of success rate and scalability, without sacrificing solution quality over randomized trials.


\section{Conclusion and Future Work}
\label{sec:conclusion}
This paper presents a scalable multi-robot motion planning algorithm combining an efficient high-level conflict resolution mechanism with MPC as the low-level planner to efficiently resolve inter-agent collision constraints through constraint splitting. We demonstrate that naively tracking the high-level plans generated from MAPF algorithms is insufficient and prone to execution failure under realistic conditions. In addition, we present results showing superior performance of CB-MPC in terms of success rate and scalability compared to other benchmarks across three different environments with varying number of robots and obstacles without compromising solution quality. Future work could include reasoning about uncertainty and/or multi-modalness in the prediction of other agent behavior using a chance-constrained framework \cite{castillo2020real} or take advantage of faster collision detection algorithms \cite{paper:zaro-collision-2023}.



\addtolength{\textheight}{-4.5cm}   




\bibliographystyle{IEEEtran}
\bibliography{references}

\begin{thebibliography}{10}
\providecommand{\url}[1]{#1}
\csname url@samestyle\endcsname
\providecommand{\newblock}{\relax}
\providecommand{\bibinfo}[2]{#2}
\providecommand{\BIBentrySTDinterwordspacing}{\spaceskip=0pt\relax}
\providecommand{\BIBentryALTinterwordstretchfactor}{4}
\providecommand{\BIBentryALTinterwordspacing}{\spaceskip=\fontdimen2\font plus
\BIBentryALTinterwordstretchfactor\fontdimen3\font minus \fontdimen4\font\relax}
\providecommand{\BIBforeignlanguage}[2]{{%
\expandafter\ifx\csname l@#1\endcsname\relax
\typeout{** WARNING: IEEEtran.bst: No hyphenation pattern has been}%
\typeout{** loaded for the language `#1'. Using the pattern for}%
\typeout{** the default language instead.}%
\else
\language=\csname l@#1\endcsname
\fi
#2}}
\providecommand{\BIBdecl}{\relax}
\BIBdecl

\bibitem{sharon2015conflict}
G.~Sharon, R.~Stern, A.~Felner, and N.~R. Sturtevant, ``Conflict-based search for optimal multi-agent pathfinding,'' \emph{Artificial Intelligence}, vol. 219, pp. 40--66, 2015.

\bibitem{andreychuk2022multi}
A.~Andreychuk, K.~Yakovlev, P.~Surynek \emph{et~al.}, ``Multi-agent pathfinding with continuous time,'' \emph{Artificial Intelligence}, p. 103662, 2022.

\bibitem{boyarski2015icbs}
E.~Boyarski, A.~Felner, R.~Stern \emph{et~al.}, ``{ICBS}: {I}mproved conflict-based search algorithm for multi-agent pathfinding,'' in \emph{International Joint Conference on Artificial Intelligence}, 2015.

\bibitem{barer2014suboptimal}
M.~Barer, G.~Sharon, R.~Stern, and A.~Felner, ``Suboptimal variants of the conflict-based search algorithm for the multi-agent pathfinding problem,'' in \emph{Seventh Annual Symposium on Combinatorial Search}, 2014.

\bibitem{li2019improved}
J.~Li, A.~Felner, E.~Boyarski \emph{et~al.}, ``Improved heuristics for multi-agent path finding with conflict-based search,'' in \emph{International Joint Conference on Artificial Intelligence}, 2019, pp. 442--449.

\bibitem{honig2016multi}
W.~H{\"o}nig, T.~S. Kumar, L.~Cohen \emph{et~al.}, ``Multi-agent path finding with kinematic constraints,'' in \emph{International Conference on Automated Planning and Scheduling}, 2016.

\bibitem{honig2019persistent}
W.~H{\"o}nig, S.~Kiesel, A.~Tinka \emph{et~al.}, ``Persistent and robust execution of mapf schedules in warehouses,'' \emph{IEEE Robotics and Automation Letters}, vol.~4, no.~2, pp. 1125--1131, 2019.

\bibitem{schneider2003potential}
F.~E. Schneider and D.~Wildermuth, ``A potential field based approach to multi robot formation navigation,'' in \emph{IEEE International Conference on Robotics, Intelligent Systems and Signal}, vol.~1, 2003, pp. 680--685.

\bibitem{tanner2005towards}
H.~G. Tanner and A.~Kumar, ``Towards decentralization of multi-robot navigation functions,'' in \emph{IEEE International Conference on Robotics and Automation}, 2005, pp. 4132--4137.

\bibitem{gayle2009multi}
R.~Gayle, W.~Moss, M.~C. Lin, and D.~Manocha, ``Multi-robot coordination using generalized social potential fields,'' in \emph{IEEE International Conference on Robotics and Automation}, 2009, pp. 106--113.

\bibitem{fox1997dynamic}
D.~Fox, W.~Burgard, and S.~Thrun, ``The dynamic window approach to collision avoidance,'' \emph{IEEE Robotics \& Automation Magazine}, vol.~4, no.~1, pp. 23--33, 1997.

\bibitem{vcap2013multi}
M.~{\v{C}}{\'a}p, P.~Nov{\'a}k, J.~Vok{\v{r}}{\'\i}nek, and M.~P{\v{e}}chou{\v{c}}ek, ``Multi-agent {RRT*}: {S}ampling-based cooperative pathfinding,'' in \emph{International Conference on Autonomous Agents and Multi-Agent Systems}, 2013, p. 1263–1264.

\bibitem{solovey2015finding}
K.~Solovey, O.~Salzman, and D.~Halperin, ``Finding a needle in an exponential haystack: Discrete {RRT} for exploration of implicit roadmaps in multi-robot motion planning,'' in \emph{Algorithmic Foundations of Robotics XI}.\hskip 1em plus 0.5em minus 0.4em\relax Springer, 2015, pp. 591--607.

\bibitem{kottinger2022conflict}
J.~Kottinger, S.~Almagor, and M.~Lahijanian, ``Conflict-based search for multi-robot motion planning with kinodynamic constraints,'' in \emph{IEEE/RSJ International Conference on Intelligent Robots and Systems}, 2022.

\bibitem{shome2020drrt}
R.~Shome, K.~Solovey, A.~Dobson \emph{et~al.}, ``{dRRT*}: {S}calable and informed asymptotically-optimal multi-robot motion planning,'' \emph{Autonomous Robots}, vol.~44, no.~3, pp. 443--467, 2020.

\bibitem{li2021optimal}
B.~Li, Y.~Ouyang, Y.~Zhang \emph{et~al.}, ``Optimal cooperative maneuver planning for multiple nonholonomic robots in a tiny environment via adaptive-scaling constrained optimization,'' \emph{IEEE Robotics and Automation Letters}, vol.~6, no.~2, pp. 1511--1518, 2021.

\bibitem{zhou2017real}
Y.~Zhou, H.~Hu, Y.~Liu \emph{et~al.}, ``A real-time and fully distributed approach to motion planning for multirobot systems,'' \emph{IEEE Transactions on Systems, Man, and Cybernetics: Systems}, vol.~49, no.~12, pp. 2636--2650, 2017.

\bibitem{chen2015decoupled}
Y.~Chen, M.~Cutler, and J.~P. How, ``Decoupled multiagent path planning via incremental sequential convex programming,'' in \emph{IEEE International Conference on Robotics and Automation}, 2015, pp. 5954--5961.

\bibitem{luis2020online}
C.~E. Luis, M.~Vukosavljev, and A.~P. Schoellig, ``Online trajectory generation with distributed model predictive control for multi-robot motion planning,'' \emph{IEEE Robotics and Automation Letters}, vol.~5, no.~2, pp. 604--611, 2020.

\bibitem{firoozi2020distributed}
R.~Firoozi, L.~Ferranti, X.~Zhang \emph{et~al.}, ``A distributed multi-robot coordination algorithm for navigation in tight environments,'' \emph{arXiv preprint arXiv:2006.11492}, 2020.

\bibitem{wagner2011m}
G.~Wagner and H.~Choset, ``M*: A complete multirobot path planning algorithm with performance bounds,'' in \emph{IEEE/RSJ international conference on intelligent robots and systems}, 2011, pp. 3260--3267.

\bibitem{arul2021v}
S.~H. Arul and D.~Manocha, ``V-rvo: Decentralized multi-agent collision avoidance using voronoi diagrams and reciprocal velocity obstacles,'' in \emph{2021 IEEE/RSJ International Conference on Intelligent Robots and Systems (IROS)}.\hskip 1em plus 0.5em minus 0.4em\relax IEEE, 2021, pp. 8097--8104.

\bibitem{li2021lifelong}
J.~Li, A.~Tinka, S.~Kiesel \emph{et~al.}, ``Lifelong multi-agent path finding in large-scale warehouses,'' in \emph{AAAI Conference on Artificial Intelligence}, vol.~35, no.~13, 2021, pp. 11\,272--11\,281.

\bibitem{velagapudi2010decentralized}
P.~Velagapudi, K.~Sycara, and P.~Scerri, ``Decentralized prioritized planning in large multirobot teams,'' in \emph{IEEE/RSJ International Conference on Intelligent Robots and Systems}, 2010, pp. 4603--4609.

\bibitem{cap2013asynchronous}
M.~C{\'a}p, P.~Nov{\'a}k, M.~Seleck{\`y} \emph{et~al.}, ``Asynchronous decentralized prioritized planning for coordination in multi-robot system,'' in \emph{IEEE/RSJ International Conference on Intelligent Robots and Systems}, 2013, pp. 3822--3829.

\bibitem{bennewitz2001optimizing}
M.~Bennewitz, W.~Burgard, and S.~Thrun, ``Optimizing schedules for prioritized path planning of multi-robot systems,'' in \emph{IEEE International Conference on Robotics and Automation}, vol.~1, 2001, pp. 271--276.

\bibitem{li2020efficient}
J.~Li, M.~Ran, and L.~Xie, ``Efficient trajectory planning for multiple non-holonomic mobile robots via prioritized trajectory optimization,'' \emph{IEEE Robotics and Automation Letters}, vol.~6, no.~2, pp. 405--412, 2020.

\bibitem{duchovn2014path}
F.~Ducho{\v{n}}, A.~Babinec, M.~Kajan \emph{et~al.}, ``Path planning with modified a star algorithm for a mobile robot,'' \emph{Procedia Engineering}, vol.~96, pp. 59--69, 2014.

\bibitem{ferranti2022distributed}
L.~Ferranti, L.~Lyons, R.~R. Negenborn \emph{et~al.}, ``Distributed nonlinear trajectory optimization for multi-robot motion planning,'' \emph{IEEE Transactions on Control Systems Technology}, 2022.

\bibitem{riegger2016centralized}
L.~Riegger, M.~Carlander, N.~Lidander \emph{et~al.}, ``Centralized {MPC} for autonomous intersection crossing,'' in \emph{IEEE International Conference on Intelligent Transportation Systems}, 2016, pp. 1372--1377.

\bibitem{malu2014kinematics}
S.~K. Malu, J.~Majumdar \emph{et~al.}, ``Kinematics, localization and control of differential drive mobile robot,'' \emph{Global Journal of Research In Engineering}, vol.~14, no.~1, pp. 1--9, 2014.

\bibitem{andersson2019casadi}
J.~A. Andersson, J.~Gillis, G.~Horn \emph{et~al.}, ``{CasADi}: {A} software framework for nonlinear optimization and optimal control,'' \emph{Mathematical Programming Computation}, vol.~11, pp. 1--36, 2019.

\bibitem{biegler2009large}
L.~T. Biegler and V.~M. Zavala, ``Large-scale nonlinear programming using {IPOPT}: {A}n integrating framework for enterprise-wide dynamic optimization,'' \emph{Computers \& Chemical Engineering}, vol.~33, no.~3, pp. 575--582, 2009.

\bibitem{castillo2020real}
M.~Castillo-Lopez, P.~Ludivig, S.~A. Sajadi-Alamdari \emph{et~al.}, ``A real-time approach for chance-constrained motion planning with dynamic obstacles,'' \emph{IEEE Robotics and Automation Letters}, vol.~5, no.~2, pp. 3620--3625, 2020.

\bibitem{paper:zaro-collision-2023}
A.~Zaro, A.~Tajbakhsh, and A.~M. Johnson, ``Collision detection for multi-robot motion planning with efficient quad-tree update and skipping,'' in \emph{arXiv:2307.07602 [cs.MA]}, 2023.

\end{thebibliography}
\end{document}